\documentclass{article}
\usepackage{spconf,amsmath,graphicx,url,times}
\usepackage{spconf,amssymb,amsmath,graphicx,times,url}
\usepackage{color}
\usepackage[lined,ruled,commentsnumbered]{algorithm2e}
\usepackage[utf8]{inputenc}
\usepackage{float}
\usepackage{comment}
\usepackage{array}
\usepackage{color}
\usepackage{soul}

\def\defequal{\stackrel{\mbox{\footnotesize def}}{=}}

\def\bal#1{\begin{align}#1\end{align}}
\newcommand{\ve}[1]{ {\mathbf{#1}} }
\newcommand{\V}{\ve{V}}
\newcommand{\W}{\ve{W}}
\renewcommand{\H}{\ve{H}}
\newcommand{\X}{\ve{X}}
\newcommand{\Y}{\ve{Y}}
\newcommand{\bPhi}{\boldsymbol{\Phi}}
\newcommand{\bphi}{\boldsymbol{\phi}}

\def\RR{\mathbb{R}}

\def\FT\text{FT}
\def\train\text{train}

\definecolor{rr}{rgb}{1,0,0} \newcommand{\rr}{\color{rr}}
\definecolor{bb}{rgb}{0,0,1} \newcommand{\bb}{\color{bb}}

\graphicspath{{Figures/}}
\DeclareGraphicsExtensions{.pdf}

\title{nonnegative matrix factorization with transform learning}
\name{Dylan Fagot, Herwig Wendt and C\'edric F\'evotte}
\address{IRIT, Universit\'e de Toulouse, CNRS, Toulouse, France\\
{\tt\small firstname.lastname@irit.fr}}

\begin{document}

\ninept
\maketitle
\sloppy

\begin{abstract}
Traditional NMF-based signal decomposition relies on the factorization of spectral data, which is typically computed by means of short-time frequency transform. In this paper we propose to relax the choice of a pre-fixed transform and learn a short-time orthogonal transform together with the factorization. To this end, we formulate a regularized optimization problem reminiscent of conventional NMF, yet with the transform as additional unknown parameters, and design a novel block-descent algorithm enabling to find stationary points of this objective function. The proposed joint transform learning and factorization approach is tested for two audio signal processing experiments, illustrating its conceptual and practical benefits. %Specifically, we obtain an improved fit between the data and its approximation and better separation performance for speech enhancement. 
\end{abstract}

\begin{keywords}
Nonnegative matrix factorization (NMF), transform learning, single-channel source separation
\end{keywords}

%%%%%%%%%%%%%%%%%%%
% BODY
%%%%%%%%%%%%%%%%%%%

%%\input{Part-Methodology.tex}
%\input{Part-Methodology-HW.tex}
%\input{Part-Experiments-Music.tex}
%\input{Part-Experiments-Source.tex}

\section{Introduction}
\label{sec:intro}

Nonnegative matrix factorization (NMF) has become a privileged approach to spectral decomposition in several fields such as remote sensing and audio signal processing. In the latter field, it has led to state-of-the-art results in source separation \cite{Smaragdis2014} or music transcription \cite{Vincent2008}. The nonnegative data $\V \in \RR_{+}^{M \times N}$ is typically the spectrogram $|\X|$ or $| \X |^{\circ 2}$ of some temporal signal $y \in \RR^{T}$, where $\X$ is a short-time frequency transform of $y$, $|\cdot|$ denotes the entry-wise absolute value and $\circ$ here denotes entry-wise exponentiation. NMF produces the approximate factorization
\bal{
\V \approx \W \H, \label{eqn:NMF}
}
where $\W \in \RR_{+}^{M \times K}$ is a nonnegative matrix referred to as {\em dictionary} that contains spectral patterns characteristic of the data while $\H \in \RR_{+}^{K \times N}$ is the nonnegative matrix that contains the {\em activation coefficients} that approximate the data samples onto the dictionary. The factorization is usually low-rank ($K < \min(M,N)$) but not necessarily so (in which case regularization constraints should apply on $\W$ and/or $\H$). The decomposition \eqref{eqn:NMF} can then be inverted back to the time domain or post-processed in various ways to solve a large panel of audio signal processing problems.

In this traditional setting, a short-time frequency transform such as the short-time Fourier, Cosine or constant-$Q$ transforms acts as a pre-processing of the raw temporal data $y$. This is a potential limitation as any ill-chosen specification of the time-frequency transform may harm the quality of the decomposition. As such, we here propose to learn the transform {\em together} with the latent factors $\W$ and $\H$. We propose to address this task by solving an optimization problem of the form
\bal{
\min_{\phi, \W, \H} D( |\phi(y)|^{\circ 2} | \W \H) \label{eqn:optim}}
subject to structure constraints on $\phi: \RR^{T} \rightarrow \RR^{M \times N} $ and to nonnegativity of $\W$ and $\H$, and where $D(\,\cdot\, | \,\cdot\,)$ is a measure of fit.
In addition, we study and promote the use of sparsity-inducing penalty on $\mathbf{H}$. We refer to objectives of the form \eqref{eqn:optim} as TL-NMF, which stands for \emph{transform-learning NMF}.\\

\noindent{\bf Connections to other works.}
TL-NMF is inspired by the work of Ravishankar \& Bresler \cite{ravishankar2013learning} on learning sparsifying transforms. Given a collection of data samples $\ve{Y}$ (such as a set of images), their work consists in finding an invertible transform $\bPhi$ such that the output of $\bPhi \ve{Y}$ is sparse. We are instead looking for a transform $\phi$ such that $| \phi(y) |^{\circ 2}$ can be well approximated by a NMF. %Note that we could more generally consider the problem of finding $\phi$ such that $\phi(y)$ is low-rank.
TL-NMF can be viewed as finding a one-layer factorizing network, where $y$ acts as the raw data, $\phi$ the linear operator, $| \cdot |^{\circ 2}$ the nonlinearity and $\W \H$ the output of the network. 
Recent work has proposed combining deep learning and NMF, but in a different way. For instance, \cite{Roux2015} considers a discriminative NMF setting and \cite{Smaragdis2017} studies nonnegative auto-encoders. The TL-NMF framework  proposed in this work could in addition be extended to fully bridge deep learning and NMF by looking for a cascade of decompositions $f_{L}(\phi_{L} \ldots f_{1}(\phi_{1} (y)))$ such that the output is a NMF. Yet, this is beyond the scope of this paper and left for future work. Moreover, TL-NMF still operates in a {\em transformed} domain and is not directly related to synthesis-based NMF models in which the raw data $y(t)$ is modeled as $y(t) = \sum_{k} c_{k}(t)$ where the spectrogram of $c_{k}(t)$ is penalized so as to be closely rank-one \cite{nips14,Kameoka2015}. \\

\noindent{\bf Goals and contributions.}
The goal of this work is to study the TL-NMF problem of form \eqref{eqn:optim}. 
As a first step, we propose in this paper to gently depart from the traditional short-time Fourier or Cosine transform setting by restricting the transform $\phi(y)$ to be a short-time orthogonal transform (Section \ref{sec:short}). We consider real-valued transforms for simplicity and use the short-time Cosine transform (STCT) as a baseline. We propose an operational algorithm that returns stationary points of~\eqref{eqn:optim} and enables to learn a structured transform $\phi$ together with the NMF factors $\mathbf{W}$ and $\mathbf{H}$ (Section~\ref{sec:algorithm}). The TL-NMF approach is put to test and compared to conventional STCT-based NMF in two benchmark audio signal processing experiments: music decomposition (Section \ref{sec:expe}) and speech enhancement (Section \ref{sec:expe2}). The results demonstrate that the proposed approach is operational and yields, in both applications, significant benefits (reduced objective function, data-adapted atoms, improved separation accuracy). 

%We present the details of our approach and its connections with the state of the art in Section~\ref{sec:learning}. Section~\ref{sec:algorithm} describes an algorithm that returns stationary points of~\eqref{eqn:optim}. Section~\ref{sec:expe} and \ref{sec:expe2} report results from music decomposition and speech enhancement experiments. In particular, we show that the proposed framework can significantly improve separation accuracy as compared to standard DCT-based NMF.

\section{NMF meets transform learning} \label{sec:learning}

\subsection{Learning a short-time orthogonal transform} \label{sec:short}

Let us denote by $\ve{Y} \in \RR^{M \times N}$ the matrix that contains adjacent and overlapping short-time frames of size $M$ of $y$ and denote by $\bPhi_{\text{DCT}} \in \RR^{M \times M}$ the orthogonal real-valued DCT matrix with coefficients $[\bPhi_{\text{DCT}}]_{qm} = (2M)^{-1/2} \cos \left(\pi (q+1/2)(m+1/2)/M \right)$.
With these notations, the power spectrogram $|\mathbf{X}|^{\circ 2}$ of $y$ is simply given by $|\bPhi_{\text{DCT}} \ve{Y} |^{\circ 2}$. 
%For the sake of generality, we use sparse-NMF  so that the problem may be cast as
Traditional NMF (with sparsity) may be cast as
%\bal{ \label{eqn:obj}
%&\min_{\W,\H} D( | \bPhi_{\text{DCT}} \ve{Y} |^{\circ 2} | \W \H )+\lambda \frac{M}{K}||\H||_1 \nonumber \\
%& \quad \text{s.t.} \ \ \W \ge 0 ,\H \ge 0,\ \ \text{and}\ \ \df{( ||\mathbf{w}_k||_1=1)_{1\leq k \leq K}}
%} 
%\bal{ \label{eqn:obj}
%\min_{\W,\H \ge 0} D( | \bPhi_{\text{DCT}} \ve{Y} |^{\circ 2} | \W \H )+\lambda \frac{M}{K}||\H||_1  \ \ \text{s.t.} \ \ \forall k, \ ||\mathbf{w}_k||_1=1
%} 
%\bal{ \label{eqn:obj}
%&\min_{\W,\H} D( | \bPhi_{\text{DCT}} \ve{Y} |^{\circ 2} | \W \H )+\lambda \frac{M}{K}||\H||_1  \nonumber \\
%& \quad \quad \quad \text{s.t.} \ \ \H \in \RR^{K \times N}_{+}, \W \in \RR^{F \times K}_{+} , \ \forall k,   ||\mathbf{w}_k||_1=1
%} 
\bal{ \label{eqn:obj}
&\min_{\W,\H} D( | \bPhi_{\text{DCT}} \ve{Y} |^{\circ 2} | \W \H )+\lambda \frac{M}{K}||\H||_1  \nonumber \\
& \quad \quad \quad \text{s.t.} \ \ \W \ge 0, \H \ge 0, \ \forall k,   ||\mathbf{w}_k||_1=1
} 
where the notation $\ve{A} \ge 0$ expresses nonnegativity, $\| \ve{A} \|_{1} = \sum_{ij} |a_{ij}|$ and $\ve{w}_{k}$ denotes the $k^{th}$ column of $\ve{W}$. Regular unpenalized NMF is simply obtained for $\lambda = 0$.
%{where $\mathbf{a}_k$ denotes the $k$-th column of $\ve{A}$, $\ve{A} \ge 0$ expresses its nonnegativity} and $\lambda=0$ corresponds to the regular NMF problem.
The normalizing ratio $M/K$ ensures the measure of fit and the penalty term are of the same order of magnitude. We propose in this work to relax the pre-fixed transform $\bPhi_{\text{DCT}}$ and learn it jointly with $\W$ and $\H$. 
In other words, we consider  the general case where $\mathbf{\Phi} \in \RR^{M \times M}$ is a parameter of the objective function defined by 
\begin{equation}
C_{\lambda}(\mathbf{\Phi},\W,\H) \defequal D( | \bPhi \ve{Y} |^{\circ 2} | \W \H )+\lambda \frac{M}{K}||\H||_1
\end{equation}
and seek a solution to the TL-NMF problem defined by
%\bal{
%\label{eqn:TLobj}
%\min_{\bPhi \in \mathbb{R}^{M \times M}, \W,\H \ge 0} C_{\lambda}(\mathbf{\Phi},\W,\H) \ \ \text{s.t.} \ \ \bPhi^{T} \bPhi = \ve{I},\ \ \W \ge 0 ,\H \ge 0, \nonumber\\
%\text{and}\ \ \df{( ||\mathbf{w}_k||_1=1)_{1\leq k \leq K}}.
%}
%\bal{
%\label{eqn:TLobj}
%&\min_{\bPhi, \W,\H} C_{\lambda}(\mathbf{\Phi},\W,\H) \ \ \text{s.t.} \ \ \bPhi^{T} \bPhi = \ve{I},\ \ \W \ge 0 ,\H \ge 0, \nonumber\\
% & \quad \quad \quad \text{s.t.} \ \ \W \in \RR^{F \times K}_{+} ,\H \in \RR^{K \times N}_{+}, \ \forall k,   ||\mathbf{w}_k||_1=1, \\
% & \quad \quad \quad \quad \quad \bPhi \in \RR^{M \times M}, \bPhi^{T} \bPhi = \ve{I}_{M}.
%}
%\bal{
%\label{eqn:TLobj}
%&\min_{\bPhi, \W,\H} C_{\lambda}(\mathbf{\Phi},\W,\H) \nonumber\\ %\ \ \text{s.t.} \ \ \bPhi^{T} \bPhi = \ve{I},\ \ \W \ge 0 ,\H \ge 0, 
% & \quad \quad \quad \text{s.t.} \ \ \H \in \RR^{K \times N}_{+}, \W \in \RR^{F \times K}_{+} , \ \forall k,   ||\mathbf{w}_k||_1=1, \nonumber \\
% & \quad \quad \quad \quad \ \ \ \bPhi \in \RR^{M \times M}, \bPhi^{T} \bPhi = \ve{I}_{M}.
%}
\bal{
\label{eqn:TLobj}
&\min_{\bPhi, \W,\H} C_{\lambda}(\mathbf{\Phi},\W,\H) \nonumber\\ %\ \ \text{s.t.} \ \ \bPhi^{T} \bPhi = \ve{I},\ \ \W \ge 0 ,\H \ge 0, 
 & \quad \quad \quad \text{s.t.} \ \ \W \ge 0, \H \ge 0, \ \forall k,   ||\mathbf{w}_k||_1=1, \bPhi^{T} \bPhi = \ve{I}_{M}.
}
We choose at this stage to impose $\bPhi$ to be orthogonal though one could consider relaxing this assumption as well. The orthogonality constraint implicitly keeps $\bPhi$ nonsingular and excludes trivial solutions such as $(\bPhi,\W,\H) =(\ve{0},\ve{0},\ve{0}) $ or $( \ve{1}_{M \times M} , \ve{1}_{M \times 1} ,   |\ve{1}_{1 \times M}\ve{Y}|^{\circ 2}) $, where $\ve{1}_{M\times N}$ denotes the $M \times N$ matrix filled with ones. In this paper, we also choose the measure of fit $D(\,\cdot\, | \,\cdot\,)$ to be the Itakura-Saito (IS) divergence $D_{\text{IS}}(\ve{A}|\ve{B}) = \sum_{ij}( a_{ij}/b_{ij} - \log (a_{ij}/b_{ij}) -1 )$. Used with power spectral data, it is known to underlie a variance-structured Gaussian composite model that is relevant to the representation of audio signals \cite{fevotte2009nonnegative} and has proven an efficient choice for audio source separation, e.g., \cite{mlsp12}. However, the proposed framework can accommodate any other measure of fit. 

\begin{algorithm}[tb]
\SetKwData{Left}{left}\SetKwData{This}{this}\SetKwData{Up}{up}
\SetKwFunction{Union}{Union}\SetKwFunction{FindCompress}{FindCompress}
\SetKwInOut{Input}{Input}\SetKwInOut{Output}{Output}
\Input{$\Y$, $\tau$, $K$ , $\lambda$}
\Output{$\bPhi$, $\W$, $\H$}
\BlankLine
Initialize $\bPhi$, $\W$ and $\H$ \\
\While{$\epsilon>\tau$}
{
$\mathbf{H}\gets\mathbf{H}\circ\left[\frac{\mathbf{W}^T\left((\mathbf{WH})^{\circ -2}\circ |\mathbf{\Phi Y}|^{\circ 2}\right) }{\mathbf{W}^T(\mathbf{WH})^{\circ -1}+ \lambda\frac{M}{K}\mathbf{1}_{K \times N}}\right]^{\frac{1}{2}}$ \\
%$\mathbf{W}\gets\mathbf{W}\circ\left[\frac{\left((\mathbf{WH})^{\circ -2}\circ |\mathbf{\Phi Y}|^{\circ %2}\right) \mathbf{H}^T}{(\mathbf{WH})^{\circ -1}\mathbf{H}^T + \lambda\frac{M}{K} \mathbf{1}_{M \times %N}\H^T}\right]^{\frac{1}{2}}$\\
$\mathbf{W}\gets\mathbf{W}\circ\left[\frac{\left((\mathbf{WH})^{\circ -2}\circ |\mathbf{\Phi Y}|^{\circ 2}\right) \mathbf{H}^T}{\left((\mathbf{WH})^{\circ -1} + \lambda\frac{M}{K} \mathbf{1}_{M \times N}\right)\H^T}\right]^{\frac{1}{2}}$\\
Normalize the columns of $\W$ so that $ \| \ve{w}_{k} \|_{1}=1$ \\
Compute $\gamma$ and $\mathbf{\Omega}$ as in Section~\ref{sec:optim} \\
$\mathbf{\Phi} \gets \pi\left(\mathbf{\Phi} + \gamma \mathbf{\Omega}\right)$ \\
Normalize $\mathbf{\Phi}$ to remove sign ambiguity\\
Compute stopping criterion $\epsilon$ as in Eq.~\eqref{eqn:stop}\\
}
\caption{TL-NMF}
\label{algorithm:algorithm1}
\end{algorithm}

\subsection{Algorithm} \label{sec:optim}
\label{sec:algorithm}

We describe a block-coordinate descent algorithm that returns stationary points of problem~\eqref{eqn:TLobj}. %Like the objective function in \eqref{eqn:obj}, 
The blocks are the individual variables $\W$, $\H$ and $\bPhi$ that are updated in turn until a convergence criterion is met. As usual of NMF problems, the objective function $C_{\lambda}\left(\mathbf{\Phi},\mathbf{W},\mathbf{H}\right)$ is nonconvex and the returned solution depends on the initialization. We use for $\W$ and $\H$ the standard multiplicative updates that can be derived from a majorization-minimization procedure  \cite{fevotte2011algorithms}. The sum-to-one constraint on the columns of $\ve{W}$ can be rigorously enforced using a change of variable, like in \cite{icassp11c,nmfvideostruct}.
%presented in, e.g., \cite{fevotte2011algorithms}, that can be derived from a majorization-minimization procedure.
%We use for $\W$ and $\H$ the standard multiplicative IS-NMF updates presented in, e.g., \cite{fevotte2011algorithms}, that can be derived from a majorization-minimization procedure. 

Let us now turn our attention towards the update of $\bPhi$. 
We propose to use a gradient-descent procedure with a line-search step selection followed by a projection onto the orthogonal constraint, following the approach of \cite{manton2002optimization}.
The main benefit of this approach is that it yields an efficient yet simple algorithm for finding a orthogonal update for $\bPhi$.
The gradient of the objective function with respect to (w.r.t.) $\mathbf{\Phi}$ can be shown to be given by
\begin{equation}
\label{eqn:gradient}
\mathbf{\nabla} \defequal \mathbf{\nabla}_{\!\mathbf{\Phi}}C_{\lambda}\left(\mathbf{\Phi},\W,\H \right) = 2\left(\mathbf{\Delta} \circ \mathbf{X}\right)\mathbf{Y}^T
\end{equation}
where $\ve{X} = \bPhi \ve{Y}$, $\mathbf{\Delta}=\hat{\mathbf{V}}^{\circ -1}-\mathbf{V}^{\circ -1}$, $\V = |\ve{X}|^{\circ 2}, \hat{\V} = \W \H$. The steepest manifold-dependent descent direction is given by the natural gradient $\mathbf{\Omega} = \mathbf{\Phi} \mathbf{\nabla}^{T} \mathbf{\Phi} - \mathbf{\nabla}$. A suitable step-size $\gamma$ is then chosen according to the Armijo rule so that the projection $\pi\left(\mathbf{\Phi}+\gamma \mathbf{\Omega}\right)$ of the updated transform onto the orthogonal constraint induces a significant decrease of the objective function \cite{manton2002optimization}.
Our block-coordinate descent algorithm is stopped when the relative variation
\bal{ \label{eqn:stop}
\epsilon^{(i)}=\frac{C_\lambda(\mathbf{\Phi}^{(i)},\mathbf{W}^{(i)},\mathbf{H}^{(i)})-C_\lambda(\mathbf{\Phi}^{(i-1)},\mathbf{W}^{(i-1)},\mathbf{H}^{(i-1)})}{C_\lambda(\mathbf{\Phi}^{(i-1)},\mathbf{W}^{(i-1)},\mathbf{H}^{(i-1)})}
} 
between iteration $i-1$ and $i$ falls below a given threshold $\tau$. The resulting TL-NMF algorithm is summarized in Algorithm \ref{algorithm:algorithm1}. The sign ambiguity on $\bPhi$ resulting from the squared magnitude operation in $|\mathbf{\Phi}\mathbf{Y}|^{\circ 2}$ is removed by imposing that the entries of the first column of $\bPhi$ are positive, likewise the DCT.
%\cf{Finally, note that our problem admits an indeterminacy on the sign of the transform elements caused by the absolute value on $|\mathbf{\Phi}\mathbf{Y}|$. To resolve it, we only consider transforms whose first column elements are positive, likewise the DCT.}
In the following experiments, we used nonnegative random values for initializing $\mathbf{W}$ and $\mathbf{H}$. 
Besides, we chose to initialize the transform $\mathbf{\Phi}$ in TL-NMF with a random orthogonal matrix.
%The transform $\bPhi$ is initialized with baseline DCT, i.e., $\bPhi = \bPhi_{\text{DCT}}$.

\begin{figure}[tb]
\centering
\begin{tabular}{cc}
$\lambda = 10^{3}$ & $\lambda = 10^{6}$ \\
\includegraphics[width=0.45\linewidth]{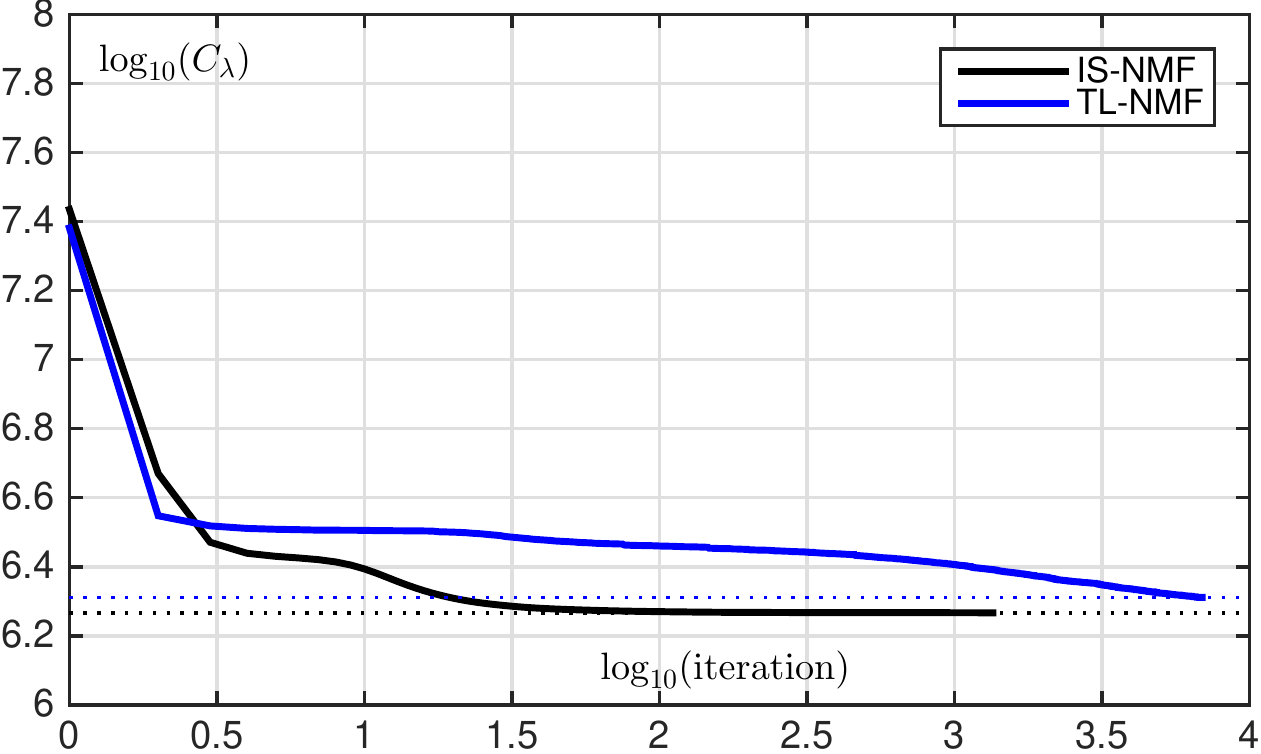} &
\includegraphics[width=0.45\linewidth]{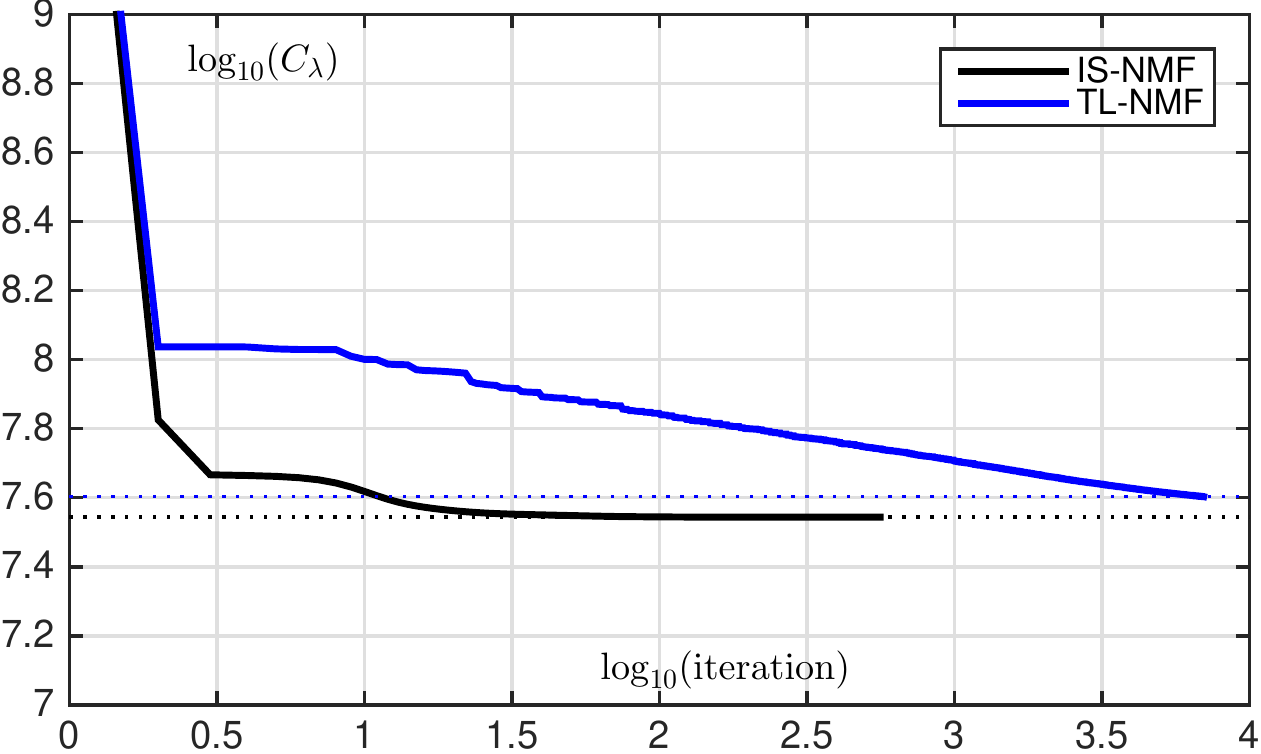}
\end{tabular}
\caption{Values of $C_{\lambda}$ w.r.t. iterations for the experiment reported in Section \ref{sec:expe}.}
\label{fig:real_div}
\end{figure}

\begin{figure}[tb]
\centering
\setlength{\tabcolsep}{0.5pt}\scriptsize
\begin{tabular}{ccc}
$\lambda=0$&$\lambda=10^3$&$\lambda=10^6$
\\
%\includegraphics[width=0.33\linewidth]{TOL1e-07_LAM1_cost}&
%\includegraphics[width=0.33\linewidth]{TOL1e-07_LAM1000_cost}&
%\includegraphics[width=0.33\linewidth]{TOL1e-07_LAM1000000_cost}%
%\\
\includegraphics[width=0.33\linewidth]{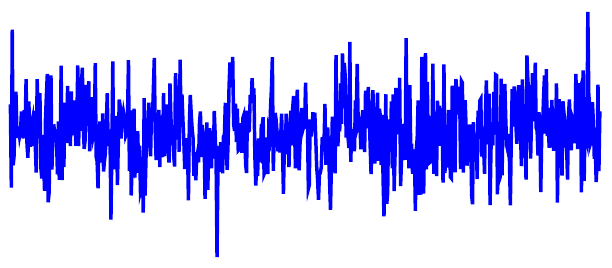}&
\includegraphics[width=0.33\linewidth]{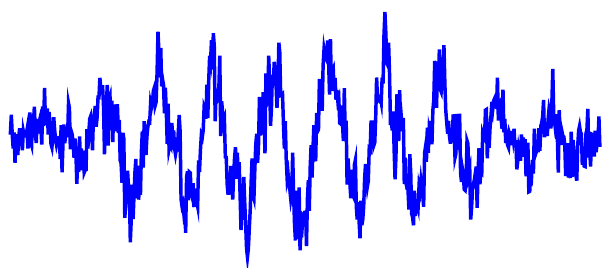}&
\includegraphics[width=0.33\linewidth]{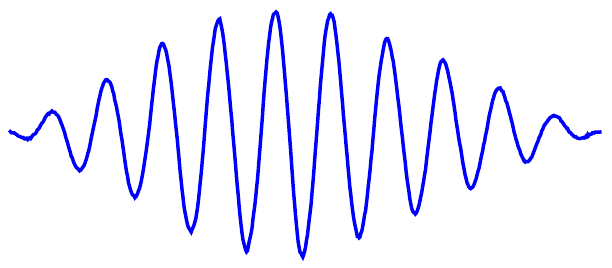}%
\vspace{-3mm}\\atom $1$\hfill\textcolor{white}{.}&&\\
\includegraphics[width=0.33\linewidth]{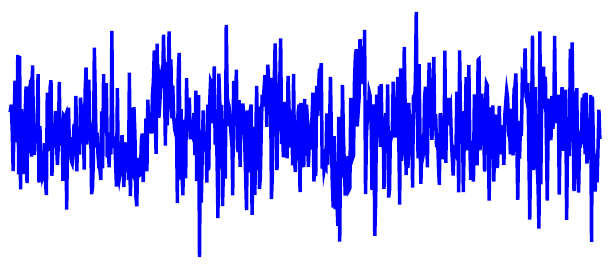}&
\includegraphics[width=0.33\linewidth]{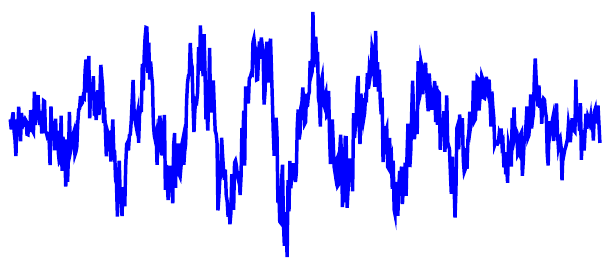}&
\includegraphics[width=0.33\linewidth]{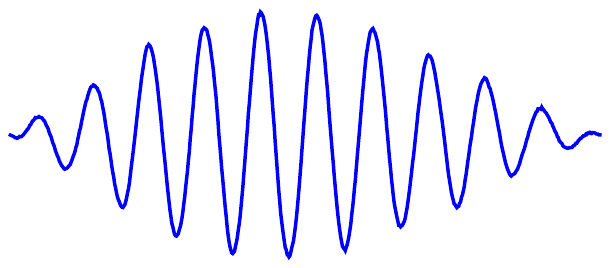}%
\vspace{-3mm}\\atom $2$\hfill\textcolor{white}{.}&&\\
\includegraphics[width=0.33\linewidth]{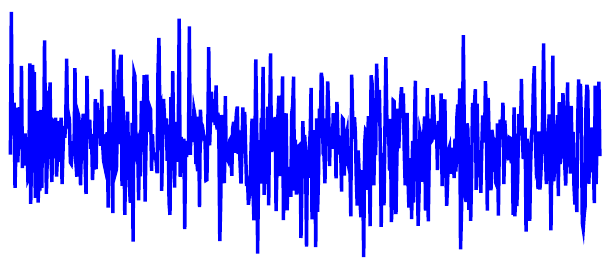}&
\includegraphics[width=0.33\linewidth]{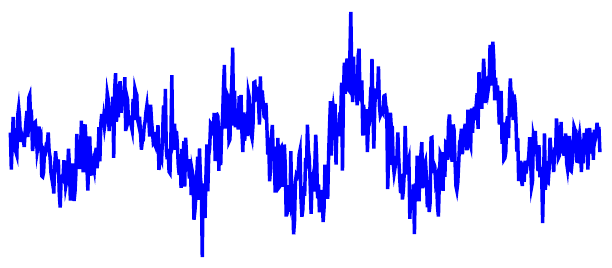}&
\includegraphics[width=0.33\linewidth]{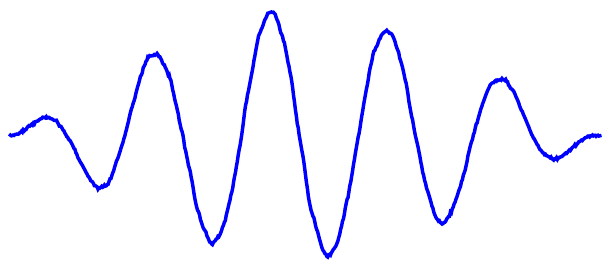}%
\vspace{-3mm}\\atom $3$\hfill\textcolor{white}{.}&&\\
\includegraphics[width=0.33\linewidth]{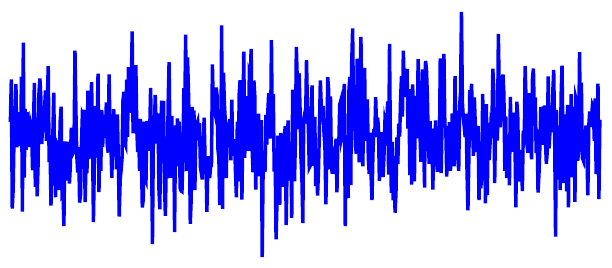}&
\includegraphics[width=0.33\linewidth]{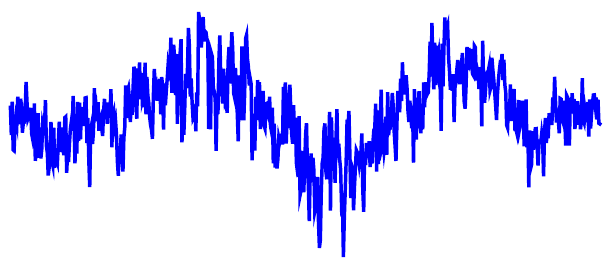}&
\includegraphics[width=0.33\linewidth]{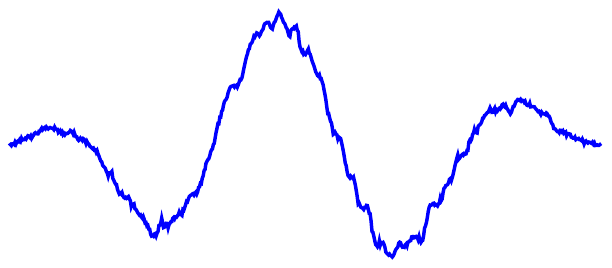}%
\vspace{-3mm}\\atom $4$\hfill\textcolor{white}{.}&&\\
\includegraphics[width=0.33\linewidth]{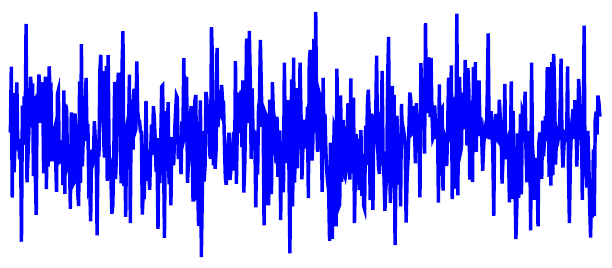}&
\includegraphics[width=0.33\linewidth]{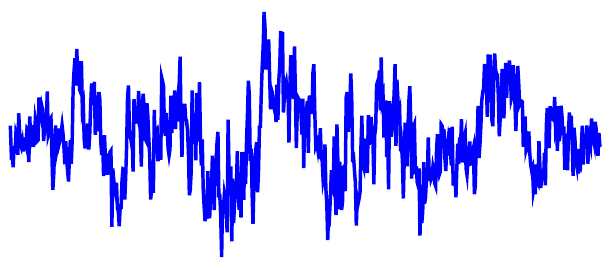}&
\includegraphics[width=0.33\linewidth]{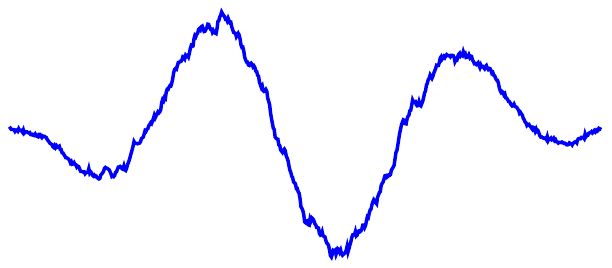}%
\vspace{-3mm}\\atom $5$\hfill\textcolor{white}{.}&&\\
\includegraphics[width=0.33\linewidth]{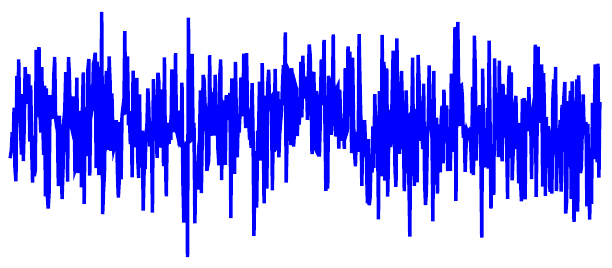}&
\includegraphics[width=0.33\linewidth]{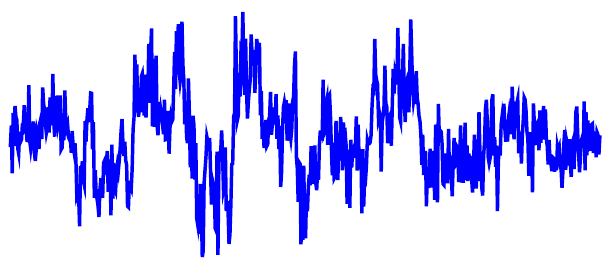}&
\includegraphics[width=0.33\linewidth]{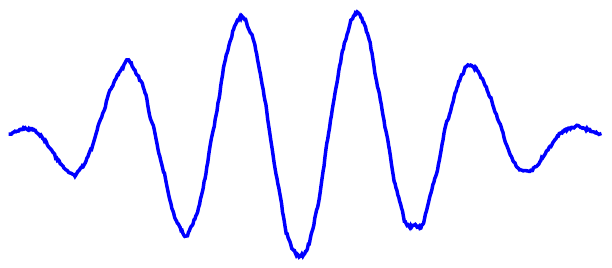}%
\vspace{-3mm}\\atom $6$\hfill\textcolor{white}{.}&&\\
\end{tabular}
\caption{\label{fig:wave} The six most significant atoms learnt by TL-NMF from random initializations for $\lambda \in\{ 0,\;10^{3}\; 10^{6}\}$ (from left to right).}
\end{figure}

\section{Experiment 1: Music decomposition} \label{sec:expe}
In this section, we report results obtained with the proposed algorithm for decomposing real audio data $y(t)$, consisting of a $23$ s excerpt of {\it Mamavatu} by Susheela Raman that has been downsampled to $f_s=16$ kHz. $\ve{Y}$ is constructed using $40$ ms-long, 50\%-overlapping temporal segments that are windowed with a sine bell. This construction leads to $M=640$ and $N=1191$.
The proposed TL-NMF is compared to conventional IS-NMF, which is obtained using Alg.~\ref{algorithm:algorithm1} as well, but fixed transform $\bPhi=\bPhi_{\text{DCT}}$.
%can be seen as special case of TL-NMF with fixed transform $\bPhi = \bPhi_{\text{DCT}$
%described by Eq.~\eqref{eqn:obj} and to Algo.~\ref{algorithm:algorithm1} with fixed transform $\bPhi=\bPhi_{\text{DCT}}$. 
The two algorithms are run with the same stopping threshold $\tau = 10^{-7}$, the same rank $K=10$, and three values of the regularization parameter $\lambda \in \{0,10^3,10^6\}$. The two algorithms are initialized with the same random initializations of $\mathbf{W}$ and $\mathbf{H}$.

\noindent{\bf Comparison of objective function values.\quad}
Fig.~\ref{fig:real_div} displays the objective function values w.r.t.~iterations for the two approaches for $\lambda=10^6$ (the initial objective value is cropped for better readability) and permits the following conclusions. 
IS-NMF reaches the stopping criterion after fewer iterations than TL-NMF, as expected since the transform $\mathbf{\Phi}$ is fixed for the latter, leading to faster convergence for the remaining variables $\mathbf{W}$ and $\mathbf{H}$. Moreover, TL-NMF starts at a larger objective function value due to the random orthogonal transform initialization for $\mathbf{\Phi}$. 
Yet, the proposed TL-NMF algorithm monotonously decreases the objective function to a value that, at convergence, is of the order of the final value of IS-NMF. %This proves the ability of TL-NMF to learn a valid transform from mere random initialization.

%is below that obtained for conventional IS-NMF.
%This indicates that TL-NMF is effective in exploiting the extra flexibility offered by learning the transform $\bPhi$ jointly with the factorization.

\noindent{\bf Regularization and learnt transform.\quad}
We now examine examples of the atoms returned by TL-NMF (rows ${{\bphi}}_m$ of $\bPhi$) for the three values of $\lambda$. Fig.~\ref{fig:wave} displays the six atoms which most contribute to the audio signal
(i.e., with largest values $||{{\bphi}}_m \Y ||_{2}$).
%in the sense that they correspond to the six largest values of $||{\bphi}_m \Y ||_{2}$.
Clearly, without regularization ($\lambda=0$), the learnt atoms lack apparent structure. Yet, interestingly, as the value for $\lambda$ is increased, the atoms become oscillatory and smoother. This is a direct consequence of and justifies the use of the sparsity-inducing term in \eqref{eqn:obj}, which induces a structure-seeking constraint on the transform. Eventually, for strong regularization ($\lambda=10^6$), the learnt atoms resemble packet-like, highly regular oscillations, further analyzed in the next paragraph.

\begin{figure}[tb]
\centering\scriptsize
\setlength{\tabcolsep}{0.5pt}
\begin{tabular}{ccc}
\includegraphics[width=0.33\linewidth]{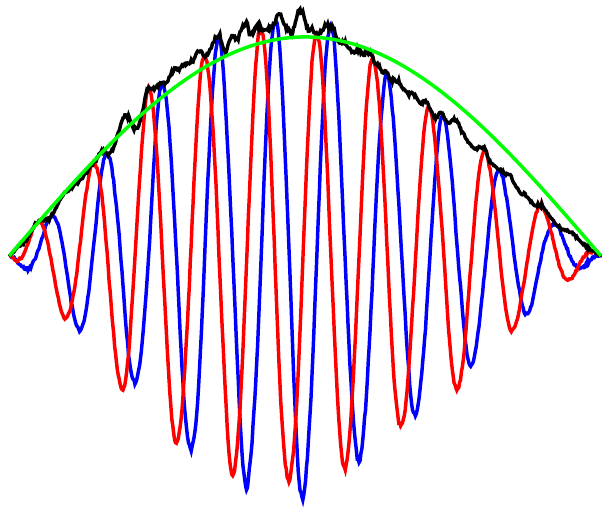}&
\includegraphics[width=0.33\linewidth]{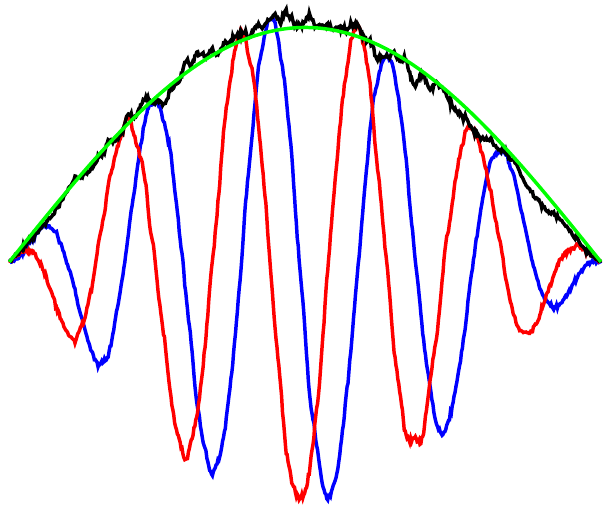}&
\includegraphics[width=0.33\linewidth]{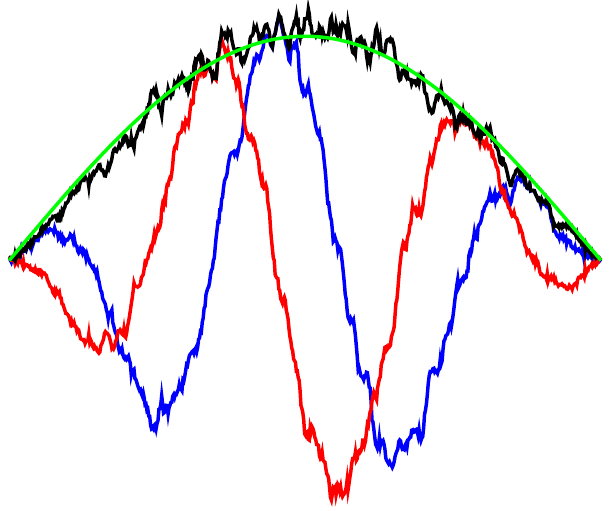}%
\vspace{-2mm}\\
atoms ${\bb1}\;\&\;{\rr2}$\hfill\textcolor{white}{.}&atoms ${\bb3}\;\&\;{\rr6}$\hfill\textcolor{white}{.}&atoms ${\bb4}\;\&\;{\rr5}$\hfill\textcolor{white}{.}
\end{tabular}
\caption{\label{fig:quadrature}
The atoms in Fig. \ref{fig:wave} for $\lambda =10^{6}$ plotted pairwise (blue and red solid lines); square root of the sum of their squared magnitude (black solid lines); sine bell window (green solid lines).}
\end{figure}

\noindent{\bf Properties of the learnt atoms.\quad}
In Fig.~\ref{fig:quadrature}, pairs of atoms $(1,2)$, $(3,6)$, $(4,5)$ (from left to right) for $\lambda=10^6$ are plotted, together with the square root of their summed square magnitudes and the sine bell used to window the data.
Interestingly, the results demonstrate that the identified pairs of atoms are approximately in \emph{quadrature}. They hence offer shift invariance properties similar to those of the (complex-valued) short time Fourier transform, which is ubiquitous  in audio applications. 
Yet, the TL-NMF algorithm permits to \emph{learn} these atoms from random initializations. This provides strong evidence for the relevance of the proposed approach.
Further, note that the atoms embrace the sine bell used to taper the observations $\mathbf{Y}$.

\section{Experiment 2: Supervised source separation}
\label{sec:expe2}

%In the previous section, we reported results of exploratory nature that show how TL-NMF is effective in learning a transform. 
We now examine whether learning an adaptive transform is actually useful for source separation. 
To this end, we consider a supervised NMF-based separation setting that follows the approach of \cite{sma07b}. In the following we address the separation of speech from interfering noise, but the method can be applied to any class of sound.

\subsection{Principle} \label{sec:princ}
We assume that we are given speech and noise training data $y_{\text{sp}}(t)$ and $y_{\text{no}}(t)$ from which we form short-time matrices $\Y_{\text{sp}}$ and $\Y_{\text{no}}$ of sizes $M \times N_{\text{sp}}$ and $M \times N_{\text{no}}$, as in Section~\ref{sec:short}. Given a noisy speech recording $y(t)$ with short-time matrix $\ve{Y}$, traditional supervised NMF amounts to estimating activation matrices $\H_{\text{sp}}$ and $\H_{\text{no}}$ such that
\bal{ \label{eqn:sss}
\V \approx \W_{\text{sp}}\H_{\text{sp}} + \W_{\text{no}} \H_{\text{no}},
}
subject to sparsity of $\H_{\text{sp}}$ and $\H_{\text{no}}$, where $\V = |\bPhi_{\text{DCT}} \ve{Y}|^{\circ 2}$, $\W_{\text{sp}} = |\bPhi_{\text{DCT}} \ve{Y}_{\text{sp}}|^{\circ 2}$, $\W_{\text{no}} = |\bPhi_{\text{DCT}} \ve{Y}_{\text{no}}|^{\circ 2}$ \cite{sma07b}. Temporal source and noise estimates are then reconstructed in a second step by so-called Wiener filtering \cite{fevotte2009nonnegative}, based on the spectrogram estimates $\hat{\V}_{\text{sp}} = \W_{\text{sp}} \H_{\text{sp}}$ and $\hat{\V}_{\text{no}} = \W_{\text{no}} \H_{\text{no}}$.

In this section, we generalize this procedure by again learning an optimal transform within the separation procedure. To this end, we propose to build an approximation like \eqref{eqn:sss} but where the fixed transform $\bPhi=\bPhi_{\text{DCT}}$ is now relaxed and learnt together with $\H_{\text{sp}}$ and $\H_{\text{no}}$.  This means we propose to minimize
\bal{
C_{\Lambda}(\bPhi, \H_{\text{sp}},\H_{\text{no}}) &\defequal D_{\text{IS}}\left(|\bPhi \Y|^{\circ 2} \, \middle |   \, {|  \bPhi \ve{Y}_{\text{sp}}|^{\circ 2}} \H_{\text{sp}} +  {|\bPhi \ve{Y}_{\text{no}}|^{\circ 2}} \H_{\text{no}}\right) \nonumber \\ 
& \quad \quad + \lambda_{\text{sp}}\frac{M}{N_{\text{sp}}} ||\H_{\text{sp}}||_1+\lambda_{\text{no}}\frac{M}{N_{\text{no}}} ||\H_{\text{no}}||_1  \nonumber \\ 
&\ \ \quad \quad \text{s.t.} \ \ \bPhi^{T} \bPhi = \ve{I},  \H_{\text{sp}} \ge 0, \H_{\text{no}} \ge 0, \label{equ:costsn}
}
with $\Lambda=\left(\lambda_{\text{sp}},\lambda_{\text{no}}\right)$ defining the possibly different weights in front of the sparsity terms. 
%which may be relevant for separating a sparse speech from a stationary noise.
%\st{The sparsity-inducing $\ell_{1}$ terms on $\H_{s}$ and $\H_{n}$ regularize the factorization which becomes potentially overcomplete in the event of large training datasets.} 
Note that $\bPhi$ now appears in both sides of the data-fitting term $D_{\text{IS}}(\cdot | \cdot)$, as the same transform is applied to the mixed data $\ve{Y}$ and the training data $\Y_{\text{sp}}$ and $\Y_{\text{no}}$. This requires to slightly modify the gradient of $C_{\Lambda}$ w.r.t. $\bPhi$ as compared to Section~\ref{sec:learning} and as described in next section. Given a solution to \eqref{equ:costsn} and $\V = | \bPhi \Y|^{\circ 2}$ along with speech and noise spectrogram estimates $\hat{\V}_{\text{sp}} = | \bPhi \ve{Y}_{\text{sp}}|^{\circ 2} \H_{\text{sp}}$ and $\hat{\V}_{\text{no}} = |  \bPhi \ve{Y}_{\text{no}}|^{\circ 2} \H_{\text{no}}$, temporal estimates may still be produced with Wiener filtering, i.e.,
\bal{
\hat{\Y}_{s} = \bPhi^{T} \left( \frac{\hat{\V}_{\text{sp}}}{\hat{\V}_{\text{sp}}+\hat{\V}_{\text{no}}} \circ (\bPhi \ve{Y}) \right)\label{eqn:wiener}
}
followed by standard overlap-adding of the columns of $\hat{\Y}_{\text{sp}}$ to return $\hat{y}_{\text{sp}}(t)$, and likewise for the noise. This is exactly the same procedure than in traditional NMF-based separation except that $\bPhi_{\text{DCT}}$ and $\bPhi_{\text{DCT}}^T$ are replaced by $\bPhi$ and $\bPhi^{T}$.

\subsection{Algorithm} \label{sec:sss}
\IncMargin{-0.5em}
Denote $\Y_{\text{tr}}=\left[\Y_{\text{sp}},\Y_{\text{no}}\right]$, $\X_{\text{tr}} = \bPhi \Y_{\text{tr}}$,  $\W=|\X_{\text{tr}}|^{\circ 2}$, $\H =\left[\H_{\text{sp}}^T , \H_{\text{no}}^T\right]^{T}$ and $\hat{\V} = \W \H$. Given $\W$, $\H$ can be updated with multiplicative rules derived from majorization-minimization as in \cite{fevotte2011algorithms}. We use again a gradient-descent approach for the update of $\bPhi$. The gradient of the objective function \eqref{equ:costsn} can be expressed as
\begin{equation}
\label{equ:grad2}
    \nabla_{\bPhi}C_{\Lambda}\left(\bPhi,\H\right)=2\left(\mathbf{\Delta} \circ \mathbf{X}\right)\Y^T + 2\left(\mathbf{\Xi} \circ \mathbf{X}_{\text{tr}}\right)\Y_{\text{tr}}^T
\end{equation}
where $\mathbf{\Delta}=\hat{\mathbf{V}}^{\circ -1}-\mathbf{V}^{\circ -1}$ and $\mathbf{\Xi}=\mathbf{\Delta}' \H^T$  with $\mathbf{\Delta}'=\frac{\hat{\mathbf{V}}-\mathbf{V}}{\hat{\mathbf{V}}^{\circ 2}}$.\\ Note that the first term of \eqref{equ:grad2} is the gradient in \eqref{eqn:gradient}. The second term is nothing but the gradient of the data-fitting term $D_{\text{IS}}$ with its first argument fixed. Based on \eqref{equ:grad2}, we again use a line-search step selection in the steepest natural gradient direction followed by a projection, like in Section~\ref{sec:optim} and following \cite{manton2002optimization}.
The resulting algorithm is summarized in Alg.~\ref{algo3}.
%  All the materials needed to run the experiments can be found online \footnote{\url{http://projectfactory.github.io}}.
\begin{algorithm}[tb]
\SetKwData{Left}{left}\SetKwData{This}{this}\SetKwData{Up}{up}
\SetKwFunction{Union}{Union}\SetKwFunction{FindCompress}{FindCompress}
\SetKwInOut{Input}{Input}\SetKwInOut{Output}{Output}
\Input{$\Y$, $\Y_{\text{tr}}$, $\tau$, $\Lambda=\left(\lambda_{\text{sp}},\lambda_{\text{no}}\right)$}
\Output{$\bPhi$, $\H$}
\BlankLine
Initialize $\bPhi$, $\H$\\
%$\mathbf{V}$, $\W$ and
\While{$\epsilon>\tau$}
{
$\mathbf{V} = \left|\mathbf{\Phi Y}\right|^{\circ 2}$, $\mathbf{W}=|\mathbf{\Phi} \mathbf{Y}_{\text{tr}}|^{\circ 2}$ \\
%$\mathbf{H}\gets \mathbf{H} \circ \left[\frac{\mathbf{W}^T\left((\mathbf{WH})^{\circ -2}\circ \mathbf{V}\right)}{\mathbf{W}^T(\mathbf{WH})^{\circ -1}+[\lambda_{\text{sp}}\frac{M}{N_{\text{sp}}}\mathbf{1}_{N \times N_{\text{sp}}},\lambda_{\text{no}}\frac{M}{N_{\text{no}}}\mathbf{1}_{N \times N_{\text{no}}}]^T}\right]^{\frac{1}{2}} $\\
$\mathbf{H}\gets \mathbf{H} \circ \left[\frac{\mathbf{W}^T\left((\mathbf{WH})^{\circ -2}\circ \mathbf{V}\right)}{\mathbf{W}^T(\mathbf{WH})^{\circ -1}+M \left[\frac{\lambda_{\text{sp}}}{N_{\text{sp}}}\mathbf{1}_{N \times N_{\text{sp}}},\frac{\lambda_{\text{no}}}{N_{\text{no}}}\mathbf{1}_{N \times N_{\text{no}}}\right]^T}\right]^{\frac{1}{2}} $\\
Compute $\gamma$ and $\mathbf{\Omega}$ as in Section~\ref{sec:sss} \\ 
$\mathbf{\Phi} \gets \pi\left(\mathbf{\Phi} + \gamma \mathbf{\Omega}\right)$\\
Normalize $\mathbf{\Phi}$ to remove sign ambiguity\\
Compute stopping criterion $\epsilon$ %as in \eqref{eqn:stop} using $C_3$
}
\caption{\label{algo3} Supervised TL-NMF}
\end{algorithm}

\subsection{Speech enhancement experiment}
\label{sec:results2}

We consider clean speech and noise data from the TIMIT corpus \cite{Garofolo1993} and the CHIME challenge,\footnote{\url{http://spandh.dcs.shef.ac.uk/chime_challenge}} respectively. For speech training data $y_{\text{sp}}(t)$, we use all utterances but the first one in the {\tt\footnotesize train/dr1/fcjf0} directory (about $21$ s in total). For noise training data $y_{\text{no}}(t)$, we use $30$ s of the file {\tt\footnotesize BGD\_150204\_010\_BUS.CH1.wav}, which contains noise recorded in a bus. A simulated mixed signal $y(t)$ of duration 3~s is generated by mixing the remaining speech utterance with another segment of the noise file (as such, the test data is not included in the training data), using signal-to-noise (SNR) ratios of -10 dB and  0 dB. The audio files' sampling frequency is $f_s=16$ kHz, and short-term matrices $\Y$, $\Y_{\text{sp}}$ and $\Y_{\text{no}}$ are constructed using $40$ ms-long, 50\%-overlapping windowed segments like in Section~\ref{sec:expe}, leading to $M=640$, $N=149$, $N_{\text{sp}}=1059$ and $N_{\text{no}}=1517$. 

Our supervised TL-NMF approach is compared to the traditional supervised NMF procedure (with the IS divergence) described in Section~\ref{sec:princ}, based on the same training data and using the same regularization parameters (only the transform $\bPhi$ differs between the two approaches). Source separation performance was assessed using the standard BSS\_eval criteria \cite{vincent2006performance}. For simplicity we set $\lambda_{\text{sp}} = \lambda_{\text{no}} = \lambda$ and tested various order of magnitudes of $\lambda$. The performance of both IS-NMF and TL-NMF were rather robust to the choice of $\lambda$, and also offered best individual performance for the same range of values. We report results with the optimal values  $\lambda=10^{-1}$ and $\lambda=10^{-4}$ for SNR = -10 dB and SNR = 0 dB, respectively. We also compute the performance criteria obtained by $\hat{y}_{\text{sp}} = \hat{y}_{\text{no}} = y/2$ as an indicative baseline. Table~\ref{tab:perf} reports the comparison results. 

%\df{The regularization parameters are respectively set to $\lambda=10^{-1}$ and $\lambda=10^{-4}$ for the two cases considered SNR$=-10$ dB and SNR$=0$ dB. Due to space limitation, we don't report the results for more different values of $\Lambda$ but these are robust for several orders of magnitude for the regularization parameter. The stopping threshold was set to $\tau = 10^{-5}$.}

The results show that the extra adaptability offered by TL-NMF is clearly beneficial as far as source separation capabilities are concerned. Indeed, TL-NMF dramatically improves the signal to distortion and interference ratios for the speech source by $8.5$, $18.5$ (SNR = -10 dB) and $4.73$, $9.05$ (SNR = 0 dB) as compared to IS-NMF. However, the signal to artifact ratios are slightly better with IS-NMF, with improvements of $-1.9$ (SNR = -10 dB) and $-1.14$ (SNR = 0 dB). Audio samples are available online.\footnote{\url{https://www.irit.fr/~Cedric.Fevotte/extras/audio_icassp2018.zip}}

%Regarding the noise source, all indexes are higher with TL-NMF.

%\textcolor{blue}{Indeed, TL-NMF improves dramaticaly the signal to distortion and interference ratios for the speech source by $8.5$, $18.5$ (-10dB) and $4.73$, $9.05$ (0dB) as compared to traditional IS-NMF. However, the signal to artifact ratio tends to be lower with gaps of $-1.9$ (-10 dB) and $-1.14$ (0dB) which is confirmed acoustically. For the noise source, all indexes are higher with TL-NMF.}

\begin{table}[tb]
\centering
\setlength{\tabcolsep}{4.5pt}
\begin{tabular}{|c||cc|cc|cc|}
\hline
   Method& \multicolumn{2}{c|}{SDR (dB)} & \multicolumn{2}{c|}{SIR (dB)} & \multicolumn{2}{c|}{SAR (dB)}\\
\hline\hline
SNR = -10 dB     & $\hat{y}_{\text{sp}}$ & $\hat{y}_{\text{no}}$ & $\hat{y}_{\text{sp}}$ & $\hat{y}_{\text{no}}$ & $\hat{y}_{\text{sp}}$ & $\hat{y}_{\text{no}}$ \\
   \hline
   Baseline & \hfill-9.50 & \hfill10.00 & \hfill-9.50 & \hfill10.00 & $\infty$ & $\infty$\\
   \hline
   IS-NMF & \hfill-6.75 & \hfill6.82 & \hfill-5.00 & \hfill\bf{13.95} & \hfill\bf{4.12} & \hfill7.93\\
   \hline
   TL-NMF & \hfill\bf{1.73} & \hfill \bf{12.29} & \hfill\bf{13.44} & \hfill13.33 & \hfill2.22 & \hfill\bf{19.20}\\
   \hline
   \hline
SNR = 0 dB     & $\hat{y}_{\text{sp}}$ & $\hat{y}_{\text{no}}$ & $\hat{y}_{\text{sp}}$ & $\hat{y}_{\text{no}}$ & $\hat{y}_{\text{sp}}$ & $\hat{y}_{\text{no}}$ \\
   \hline
   Baseline & \hfill0.10 & \hfill0.08 & \hfill0.10 & \hfill0.08 & $\infty$ & $\infty$\\
   \hline
   IS-NMF & \hfill1.73 & \hfill0.69 & \hfill3.06 & \hfill5.32 & \hfill\bf{9.30} & \hfill3.65\\
   \hline
   TL-NMF & \hfill\bf{6.50} & \hfill\bf{5.81} & \hfill\bf{12.11} & \hfill\bf{9.16} & \hfill8.16 & \hfill\bf{9.00}\\
   \hline
\end{tabular}
\caption{\label{tab:perf}Source separation performance.}\vspace{-3mm}
\end{table}

%Fig.~\ref{fig:cprime} displays the values of the objective function $C_{e}$ returned by supervised TL-NMF and supervised IS-NMF (in which case $\bPhi = \bPhi_{\text{DCT}}$). It clearly indicates that, at convergence, the value of the objective function obtained by the proposed algorithm is nearly one order of magnitude lower than that of IS-NMF: the latter algorithm makes the objective function reach a value of $9.5\times 10^4$ (IS divergence of $6.8\times 10^4$) while our algorithm brings the objective function value down to $1.5 \times 10^4$ (IS divergence of $6.0 \times 10^3$).
%
%\begin{figure}[t]
%  \centering
%    \centerline{\includegraphics[width=1\linewidth]{Figures/separation_panorama_u-eps-converted-to.pdf}}
%  \caption{
% Values of $C_{e}$ w.r.t. iterations for the experiment reported in Section \ref{sec:results2} (log-log scale).   }
%  \label{fig:cprime}
%\end{figure}

\section{Conclusion and future work}
\label{sec:conclusion}

We addressed the task of learning the transform underlying NMF-based signal decomposition jointly with the factorization. Specifically, we have proposed a block-coordinate descent algorithm that enables us to find a orthogonal transform $\bPhi$ jointly with the dictionary $\W$ and the activation matrix $\H$. To our knowledge, the proposed algorithm is the first operational procedure for learning a transform in the context of NMF. 
Our preliminary experiments with real audio data indicate that automatically adapting the transform to the signal pays off when seeking latent factors that accurately represent the data. In particular, we were able to retrieve a naturally structured transform which embeds some form of invariance in a first musical example. Then, we achieved source separation performance that compares very favorably against the state-of-the-art in a speech enhancement experiment.
% the improvement in data fit permits to achieve source separation performance that compares very favorably against the state-of-the-art. 
Note that although our presentation focused on the processing of audio data, the approach can be adapted to many other settings where NMF is applied to preprocessed data.

Future work will include the estimation of complex-valued transforms which can be directly compared to the STFT, the influence of the initialization of $\bPhi$, the influence of the value of $K$ on the learnt transform as well as relaxations of the orthogonal constraint on $\bPhi$ to mere nonsingular matrices.
Also, the use of alternative optimization strategies that lend themselves well to dealing with nonconvex problems in high dimension, including stochastic gradient descent, will be investigated.


\newpage

\begin{thebibliography}{10}

\bibitem{Smaragdis2014}
P.~Smaragdis, C.~F{\'e}votte, G.~Mysore, N.~Mohammadiha, and M.~Hoffman,
  ``Static and dynamic source separation using nonnegative factorizations: {A}
  unified view,'' \emph{IEEE Signal Processing Magazine}, vol.~31, no.~3, pp.
  66--75, May 2014.
  
\bibitem{Vincent2008}
E.~Vincent, N.~Bertin, and R.~Badeau, ``Harmonic and inharmonic nonnegative
  matrix factorization for polyphonic pitch transcription,'' in
  \emph{Proc.~IEEE International Conference on Acoustics, Speech and Signal
  Processing (ICASSP)}, 2008.

\bibitem{fevotte2009nonnegative}
C.~F{\'e}votte, N.~Bertin, and J.-L. Durrieu, ``Nonnegative matrix
  factorization with the itakura-saito divergence : With application to music
  analysis,'' \emph{Neural Computation}, vol.~21, no.~3, pp. 793--830, 2009.

\bibitem{mlsp12}
B.~King, C.~F{\'e}votte, and P.~Smaragdis, ``Optimal cost function and
  magnitude power for {NMF}-based speech separation and music interpolation,''
  in \emph{Proc.~IEEE International Workshop on Machine Learning for Signal
  Processing (MLSP)}, 2012.

\bibitem{ravishankar2013learning}
S.~Ravishankar and Y.~Bresler, ``Learning sparsifying transforms,'' \emph{IEEE
  Transactions on Signal Processing}, vol.~61, no.~5, pp. 1072--1086, 2013.

\bibitem{Roux2015}
J.~L. Roux, J.~R. Hershey, and F.~Weninger, ``Deep NMF for speech separation,''
  in \emph{Proc.~IEEE International Conference on Acoustics, Speech and Signal
  Processing (ICASSP)}, 2015.

\bibitem{Smaragdis2017}
P.~Smaragdis and S.~Venkataramani, ``A neural network alternative to
  non-negative audio models,'' in \emph{Proc.~IEEE International Conference on
  Acoustics, Speech and Signal Processing (ICASSP)}, 2017.

%\bibitem{shrikant2017endtoend}
%S.~Venkataramani and P.~Smaragdis, ``End-to-end Source Separation with Adaptive Front-Ends,'' arXiv preprint arXiv:1705.02514, 2017.

\bibitem{nips14}
C.~F\'evotte and M.~Kowalski, ``Low-rank time-frequency synthesis,'' in
  \emph{Advances in Neural Information Processing Systems (NIPS)}, 2014.

\bibitem{Kameoka2015}
H.~Kameoka, ``Multi-resolution signal decomposition with time-domain
  spectrogram factorization,'' in \emph{Proc.~IEEE International Conference on
  Acoustics, Speech and Signal Processing (ICASSP)}, 2015.

%\bibitem{Huang2014}
%P.-S.~Huang, M.~Kim, M.~Hasegawa-Johnson and P.~Smaragdis, ``Deep Learning for Monaural Speech Separation,'' in \emph{Proc.~IEEE International Conference on Acoustics, Speech and Signal Processing (ICASSP)}, 2014.

\bibitem{fevotte2011algorithms}
C.~F{\'e}votte and J.~Idier, ``Algorithms for nonnegative matrix factorization
  with the $\beta$-divergence,'' \emph{Neural Computation}, vol.~23, no.~9, pp.
  2421--2456, 2011.

\bibitem{icassp11c}
A.~Lef\`evre, F.~Bach, and C.~F\'evotte.
\newblock {I}takura-{S}aito nonnegative matrix factorization with group
  sparsity.
\newblock In {\em Proc.~IEEE International Conference on Acoustics, Speech and
  Signal Processing (ICASSP)}, 2011.

\bibitem{nmfvideostruct}
S.~Essid and C.~F{\'e}votte.
\newblock Smooth nonnegative matrix factorization for unsupervised audiovisual
  document structuring.
\newblock {\em IEEE Transactions on Multimedia}, vol.~15, no.~2, pp. 415--425, 2013.

\bibitem{manton2002optimization}
J.~H. Manton, ``Optimization algorithms exploiting unitary constraints,''
  \emph{IEEE Transactions on Signal Processing}, vol.~50, no.~3, pp. 635--650,
  2002.

\bibitem{sma07b}
P.~Smaragdis, B.~Raj, and M.~V. Shashanka, ``Supervised and semi-supervised
  separation of sounds from single-channel mixtures,'' in \emph{Proc.~International Conference on Independent Component Analysis and Signal
  Separation (ICA)}, 2007.

\bibitem{Garofolo1993}
J.~S. Garofolo, L.~F. Lamel, W.~M. Fisher, J.~G. Fiscus, and D.~S. Pallett,
  ``Timit acoustic-phonetic continuous speech corpus LDC93S1,'' Philadelphia:
  Linguistic Data Consortium, Tech. Rep., 1993.

\bibitem{vincent2006performance}
E.~Vincent, R.~Gribonval, and C.~F{\'e}votte, ``Performance measurement in
  blind audio source separation,'' \emph{IEEE Transactions on Audio, Speech,
  and Language Processing}, vol.~14, no.~4, pp. 1462--1469, 2006.


%\vskip5mm
%\hw{\bf ADD RELEVANT REFERENCES!}

\end{thebibliography}
\end{document}